\UseRawInputEncoding
\documentclass[letterpaper, 10 pt, conference]{ieeeconf}  
\makeatletter
\def\endfigure{\end@float}
\makeatother
\IEEEoverridecommandlockouts                              
\usepackage{amsmath,amsfonts}
\usepackage{algorithmic}
\usepackage{algorithm}
\usepackage{array}
\usepackage[caption=false,font=normalsize,labelfont=sf,textfont=sf]{subfig}
\usepackage{textcomp}
\usepackage{stfloats}
\usepackage{url}
\usepackage{verbatim}
\usepackage{graphicx}
\usepackage{cite}
\usepackage{multirow} 
\usepackage{tabularx}
\usepackage{booktabs}
\usepackage{amssymb}
\usepackage{pifont}
\usepackage{pifont}
\newcommand{\cmark}{\ding{51}}

\usepackage{times}
\usepackage{listings}
\usepackage{xcolor}
\usepackage{hyperref}
\hypersetup{
    colorlinks=true,
    urlcolor=blue, 
    linkcolor=black,
    citecolor=black
}
\lstdefinestyle{jsonstyle}{
    language=Python,
    basicstyle=\ttfamily\small,
    keywordstyle=\color{blue},
    stringstyle=\color{black},
    showstringspaces=false,
    breaklines=true,
    frame=single,
    backgroundcolor=\color{gray!10},
    morekeywords={name,dependencies,clip_keywords}
}

\title{\LARGE \bf
DART-LLM: Dependency-Aware Multi-Robot Task Decomposition and Execution using Large Language Models
}

\author{Yongdong Wang$^{1}$, Runze Xiao$^{1}$, Jun Younes Louhi Kasahara$^{1}$,\\  Ryosuke Yajima $^{1}$, Keiji Nagatani$^{2}$, Atsushi Yamashita$^{3}$, and Hajime Asama$^{4}$
\thanks{$^{1}$Yongdong Wang, Runze Xiao, Jun Younes Louhi Kasahara and Ryosuke Yajima are with Graduate School of Engineering, The University of Tokyo, Tokyo 113-8656, Japan}%
\thanks{$^{2}$Keiji Nagatani is with Faculty of Systems and Information Engineering, The University of Tsukuba, Ibaraki 305-0006, Japan}%
\thanks{$^{3}$Atsushi Yamashita is with Graduate School of Frontier Sciences, The University of Tokyo, Tokyo 277-8563, Japan}%
\thanks{$^{4}$Hajime Asama is with the Tokyo College, The University of Tokyo, Tokyo 113-0033, Japan}%
\thanks{\raggedright Project Webpage: \href{https://wyd0817.github.io/project-dart-llm/}{project-dart-llm}}
\thanks{Correspondence to 
        {\tt\small wangyongdong@robot.t.u-tokyo.ac.jp}}%
}

\begin{document}

\maketitle
\thispagestyle{empty}
\pagestyle{empty}

\begin{abstract}
Large Language Models (LLMs) have demonstrated promising reasoning capabilities in robotics; however, their application in multi-robot systems remains limited, particularly in handling task dependencies. This paper introduces DART-LLM, a novel framework that employs Directed Acyclic Graphs (DAGs) to model task dependencies, enabling the decomposition of natural language instructions into well-coordinated subtasks for multi-robot execution. DART-LLM comprises four key components: a Question-Answering (QA) LLM module for dependency-aware task decomposition, a Breakdown Function module for robot assignment, an Actuation module for execution, and a Vision-Language Model (VLM)-based object detector for environmental perception, achieving end-to-end task execution. Experimental results across three task complexity levels demonstrate that DART-LLM achieves state-of-the-art performance, significantly outperforming the baseline across all evaluation metrics. Among the tested models, DeepSeek-r1-671B achieves the highest success rate, whereas Llama-3.1-8B exhibits superior response time reliability. Ablation studies further confirm that explicit dependency modeling notably enhances the performance of smaller models, facilitating efficient deployment on resource-constrained platforms.
Please refer to the project website \url{https://wyd0817.github.io/project-dart-llm/} for videos and code.

\end{abstract}

\section{Introduction}
\label{sec:Introduction}
Large Language Models (LLMs) have demonstrated significant reasoning capabilities that can be applied to robot systems.
Chen et al. introduced the Decision Transformer, a reinforcement learning method that redefines decision-making processes as sequence modeling problems \cite{chen2021decision}. 
%
%
%
Reed et al. developed Gato, a generalist artificial intelligence model capable of learning and adapting across multiple tasks and modalities \cite{reed2022generalist}. 
%
%
%
Ahn et al.'s FLAN-SayCan and PaLM-SayCan leverage the capabilities of large pre-trained language models (LLMs) to interpret and execute tasks based on natural language instructions, enabling robots to understand and perform tasks driven by complex natural language commands \cite{ahn2022can}. 
%
%
%
Huang et al.'s Inner Monologue method integrates language models into planning and reasoning processes, enhancing robots' decision-making capabilities \cite{huang2022inner}. 
%
%
%
Brohan et al. proposed RT-1, a model for real-time control and decision-making in complex environments, applying transformer architecture directly to robotics technology to ensure unprecedented precision and adaptability in understanding and navigating the physical world \cite{brohan2022rt}.
Brohan et al. further introduced the RT-2 model, which integrates vision, language, and action capabilities, enabling robots to autonomously learn and control actual actions from web knowledge \cite{brohan2023rt}. 
Driess et al.'s Palm-E model enhances the perception and action capabilities of language models, offering a new way for artificial intelligence systems to interact more naturally with the external world through enhanced multimodal understanding \cite{driess2023palm}. 
Liang et al. launched Code as Policies, enabling large language models (LLMs) to convert natural language commands into robot strategy code with minimal prompts \cite{liang2023code}.
%
%
%
Huang et al. introduced VoxPoser, which extracts language-conditioned affordances and constraints from LLMs and applies them to the perceptual space through VLMs. VoxPoser uses a code interface without additional training for any component \cite{huang2023voxposer}.

Despite significant advancements in LLM-powered robotics research, most studies have primarily focused on single-robot systems, while applications in multi-robot systems remain at an early stage of exploration. Multi-robot systems demonstrate significant potential in complex task scenarios such as construction and disaster rescue, improving task execution efficiency and enabling collaborative problem-solving for tasks that are difficult for a single robot to accomplish independently \cite{nagatani2021innovative}. However, multi-robot task planning still faces numerous challenges, particularly concerning the complexity of mobile robots and real-time execution.
The RoCo framework proposed by Zhao et al. employs LLMs for high-level task communication and integrates multi-arm motion planning to accelerate trajectory generation. It also incorporates environmental feedback (e.g., collision detection), allowing the LLM to dynamically adjust task plans based on context. However, RoCo is designed for fixed-position robotic arm systems, making its task planning approach less adaptable to mobile robots or broader task scenarios \cite{mandi2023roco}.
Kannan et al.’s SMART-LLM has made progress in multi-robot task planning. However, it generates outputs using Python and does not explicitly handle dependencies among complex tasks. Due to the inherent complexity and rigidity of programming languages, practical testing revealed a high failure rate. Furthermore, SMART-LLM requires generating Python code that must be manually executed, lacking real-time LLM inference capabilities \cite{kannan2023smart}.

\begin{figure*}[tb]
	\centering
	\includegraphics[width=\linewidth]{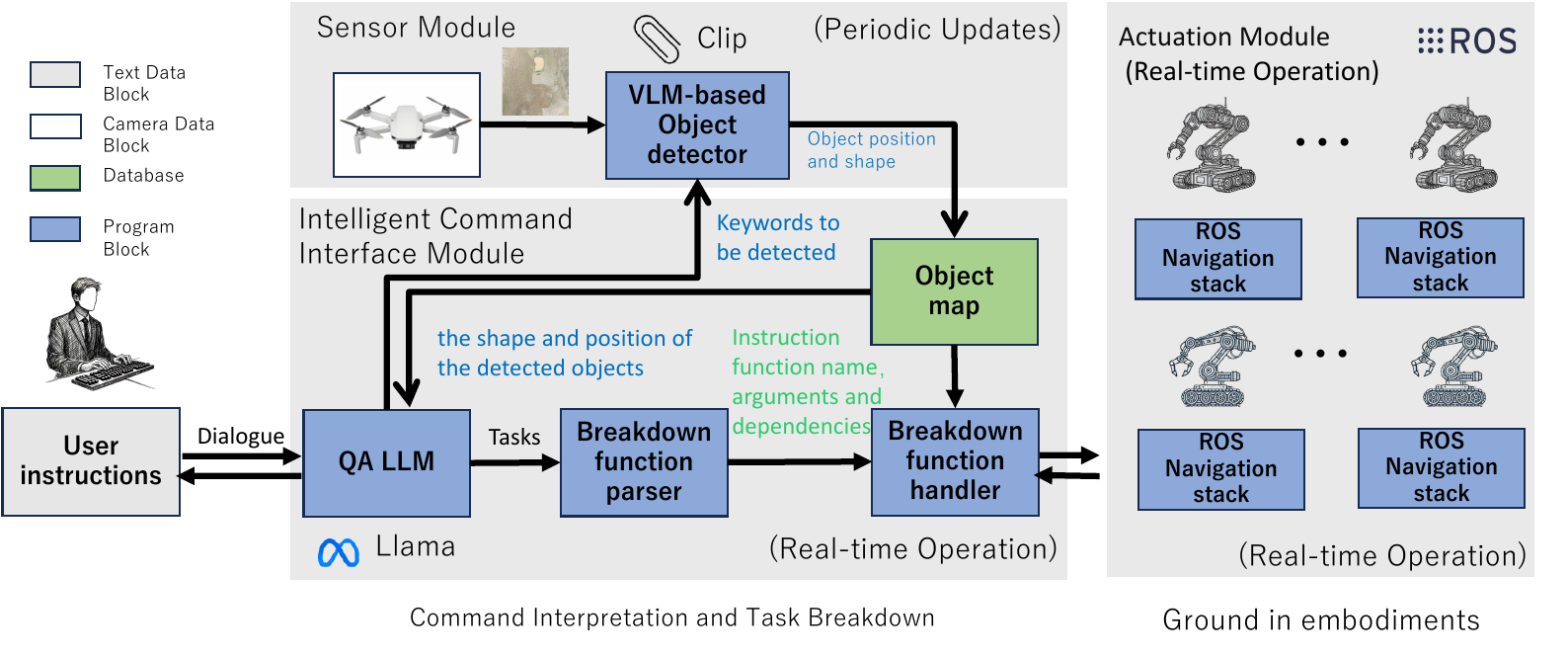}
                \caption{An overview of the DART-LLM system architecture. The system is divided into three main modules: the Sensor Module, the Intelligent Command Interface Module, and the Actuation Module. The Sensor Module captures and processes data using a Vision-Language Model (VLM)-based object detector, updating the object map with detected items. The Intelligent Command Interface Module interprets user instructions via a Question-Answering Large Language Model (QA LLM), decomposing tasks into subtasks with dependencies through the Breakdown Function Parser and Handler. This allows the establishment of complex task dependencies and coordination between multiple robots. Finally, the Actuation Module executes real-time operations using the ROS Navigation stack, guiding each robot according to the parsed and dependency-aware instructions.}
	\label{Fig:overview}
\end{figure*}

The comparison of the above-mentioned studies is shown in Table~\ref{Tab:RelatedWorkComparison}. 
Table~\ref{Tab:RelatedWorkComparison} highlights two key challenges that persist in multi-robot systems: (1) existing systems lack support for end-to-end real-time execution across multiple types of mobile robots; (2) current methods rely solely on the model’s intrinsic reasoning ability without explicitly representing task dependencies, which results in suboptimal performance, particularly for smaller models in handling complex tasks. 
For instance, RoCo \cite{mandi2023roco} lacks support for mobile robots, being designed for fixed-position robotic arms, while SMART-LLM \cite{kannan2023smart} does not support end-to-end real-time execution and fails to explicitly model task dependencies.
To address these limitations, we propose DART-LLM, A Dependency-Aware Task Decomposition and Execution System for Multi-Robot Coordination, as illustrated in Fig.~\ref{Fig:overview}.
This study makes the following contributions:

1. Dependency-Aware Task Decomposition Mechanism: A novel task decomposition mechanism that utilizes a Directed Acyclic Graph (DAG) to model subtask dependencies. This method significantly enhances system reasoning capabilities, particularly improving the ability of smaller models to handle complex task dependencies. It ensures proper task execution order while maximizing overall efficiency.

2. End-to-End Real-Time Execution Framework: A real-time execution framework includes the QA LLM module, Breakdown Function module, Actuation module, and a Vision-Language Model (VLM)-based object detection module. 
These modules perform Instruction Parsing and Task Decomposition, Parsing of Decomposed Tasks, Grounding in Embodiments, and Updating the Object Map Database to accomplish task decomposition and execution from natural language instructions to robotic actions.

3. Benchmark Dataset for Construction Robot Evaluation: A dataset of 102 high-level natural language instructions for construction robot tasks, designed with strict execution order constraints and spanning three complexity levels.

\begin{figure*}[tb]
	\centering
	\includegraphics[width=0.8\linewidth]{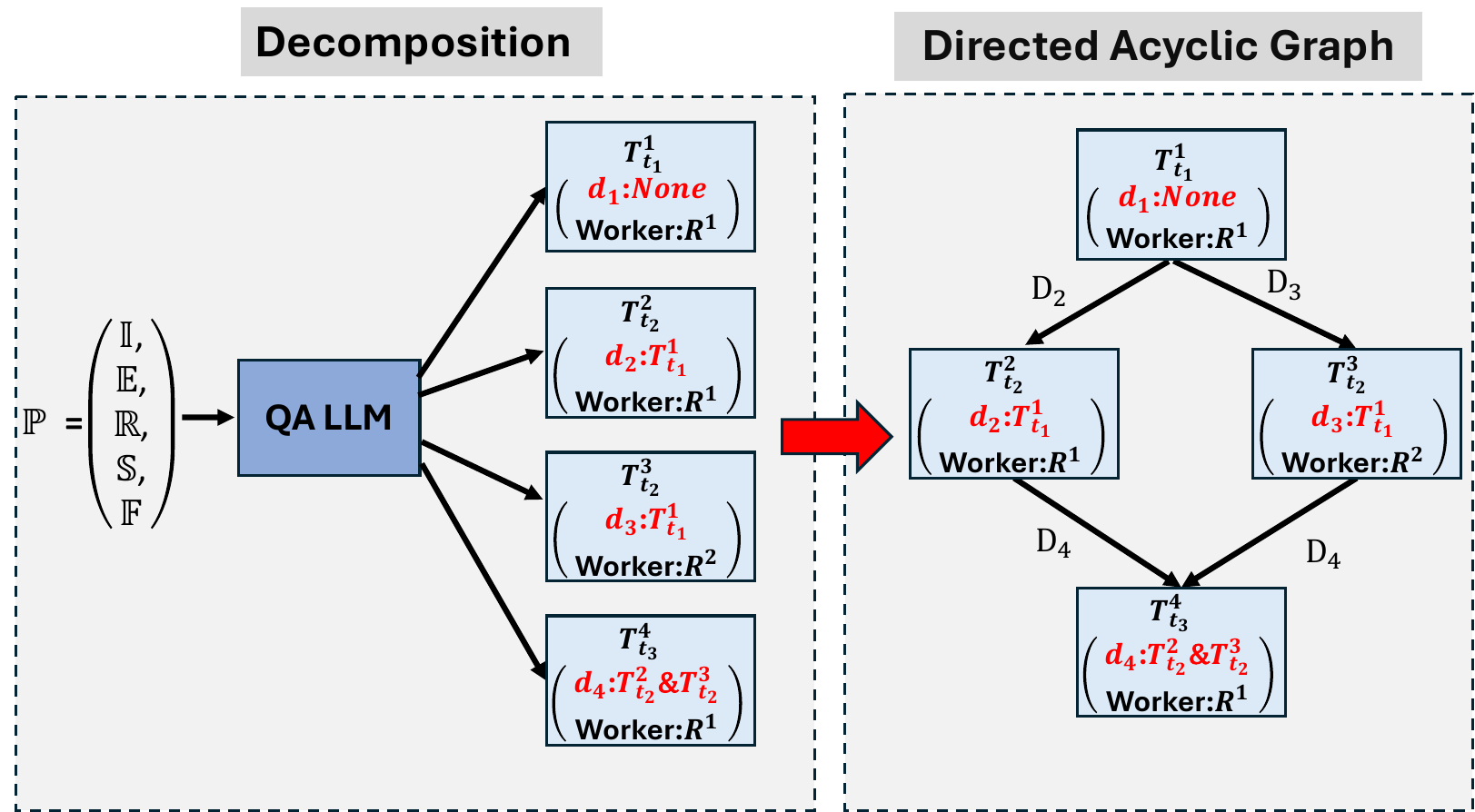}
                \caption{Dependency-aware task decomposition in DART-LLM. Left: QA LLM decomposes high-level instruction into subtasks with explicit dependency lists ($d_k$) and assigned workers. Right: Construction of the Directed Acyclic Graph (DAG) based on dependency relationships, establishing the execution order where $T^1_{t_1}$ is executed first, followed by parallel execution of $T^2_{t_2}$ and $T^3_{t_2}$, and finally $T^4_{t_3}$ after its dependencies are satisfied.}
	\label{Fig:dag_example}
\end{figure*}

\begin{table*}[tb]
\centering
\small
\caption{Comparison of Related Work in Multi-Robot Systems and Large Language Models (LLMs)}
\label{Tab:RelatedWorkComparison}
\resizebox{0.8\textwidth}{!}{%
\begin{tabular}{@{}lcccc@{}}
\toprule
\multicolumn{1}{c}{\textbf{Related Work}} & 
\textbf{\scriptsize Multi-Robot System} & 
\textbf{\scriptsize Robot Mobility} & 
\textbf{\scriptsize Real-time Capability} & 
\textbf{\scriptsize Dependency-Aware} \\
\midrule
\scriptsize Chen et al., Decision Transformer \cite{chen2021decision} & & \cmark & \cmark & \\
\scriptsize Reed et al., Gato \cite{reed2022generalist} & & \cmark & \cmark & \\
\scriptsize Ahn et al., FLAN-SayCan, PaLM-SayCan \cite{ahn2022can} & & \cmark & \cmark & \\
\scriptsize Huang et al., Inner Monologue \cite{huang2022inner} & & \cmark & \cmark & \\
\scriptsize Brohan et al., RT-1 \cite{brohan2022rt} & & \cmark & \cmark & \\
\scriptsize Brohan et al., RT-2 \cite{brohan2023rt} & & \cmark & \cmark & \\
\scriptsize Driess et al., Palm-E \cite{driess2023palm} & & \cmark & \cmark & \\
\scriptsize Liang et al., Code as Policies \cite{liang2023code} & & \cmark & & \\
\scriptsize Huang et al., VoxPoser \cite{huang2023voxposer} & & & \cmark & \\
\scriptsize Zhao et al., RoCo \cite{mandi2023roco} & \cmark & & \cmark & \\
\scriptsize Kannan et al., SMART-LLM \cite{kannan2023smart} & \cmark & \cmark & & \\
\scriptsize \textbf{Proposed DART-LLM} & \cmark & \cmark & \cmark & \cmark \\
\bottomrule
\end{tabular}%
}
\end{table*}

\section{Problem Formulation}
\label{sec:Problem_Formulation}
This section presents the foundational mathematical concepts of the DART-LLM system. We formalize the definitions of robot sets and their associated skills, team formation with collaborative skill enhancement, task decomposition into sub-tasks with temporal and dependency relationships, and the skill library essential for task execution.

\subsection{Robot Set and Skills}
We define the set of all robots as:
\begin{equation}
\mathbb{R} = \{R^1, R^2, \ldots, R^N\}
\end{equation}
where \( N \) is the number of robots. The set of skills for all robots are:
\begin{equation}
\mathbb{S} = \{S^1, S^2, \ldots, S^N\}
\end{equation}

\subsection{Team of Robots and Team Skills}
A team of robots is defined as:
\begin{equation}
\mathbb{A} = \{A^1, A^2, \ldots, A^Q\}
\end{equation}
where \( Q \) denotes the number of robots in the team. The skills of a team of robots are expressed as:
\begin{equation}
\mathbb{S}_A = \{S_A^1, S_A^2, \ldots, S_A^\Omega\}
\end{equation}
where \(\Omega \geq Q\). Here, \(\Omega\) represents the total number of skills, including both the individual skills of each robot and the new skills generated through robot collaboration. The set of all skills across all teams formed by \( N \) robots is:
\begin{equation}
\Delta = \bigcup \mathbb{S}_A
\end{equation}
It should be noted that \(\Delta\) includes \(\mathbb{S}\), the set of all individual robot skills.

\subsection{Task and Function Requirements}

The high-level language instruction is represented as \(\mathbb{I}\), and the environment is denoted by \(\mathbb{E}\). The instruction \(\mathbb{I}\) is decomposed using the \textit{QA LLM} module into a set of sub-tasks. A set of sub-tasks is defined as:
\begin{equation}
\mathbb{T} = \{T^{1}_{t_1}, T^{1}_{t_2}, \ldots, T^{K}_{t_j}\}
\end{equation}
where \( K \) is the number of sub-tasks, and \( t_j \) denotes the temporal order of a sub-task. Since sub-tasks may be executed simultaneously, we have \( j \leq K \). Each task \( T^{k}_{t_j} \) should be executable using the skills from the skill library within \(\Delta\). 

\subsection{Dependency Relationships Between Tasks}
Complex tasks in multi-robot systems often involve multiple subtasks with interdependencies. These dependencies are managed through task decomposition using a Directed Acyclic Graph (DAG).
The DAG is defined as \( G = (\mathbb{T} , \mathbb{D}) \), where:
\begin{equation}
\mathbb{D} \subseteq \mathbb{T}  \times \mathbb{T} 
\end{equation}
represents the set of directed edges indicating dependencies between subtasks. An edge \( \mathbb{D}_k = (T^{k-1}_{t_i}, T^{k}_{t_j}) \in \mathbb{D} \) indicates that subtask \( T^{k}_{t_j} \) depends on the successful completion of subtask \( T^{k-1}_{t_i} \).

\section{The DART-LLM Framework}
\label{sec: DART_LLM_Framework}
This section provides an overview of DART-LLM. The overall control architecture of DART-LLM is illustrated in Fig.~\ref{Fig:overview}. Communication between all modules is based on Robot Operating System 2 (ROS2) topics \cite{Steven2022ROS2}.

To illustrate the workflow of DART-LLM, Algorithm 1 presents the pseudocode for the system's dependency-aware multi-robot task decomposition and execution process. This algorithm outlines the system's processing of natural language instructions through multiple stages, including instruction parsing, dependency-aware multi-robot task decomposition, dependency-aware task parsing and robot assignment, and task execution with actuation. Following this high-level overview, we detail each module shown in Fig.~\ref{Fig:overview} and explain how they implement the different stages of Algorithm 1 in the subsequent subsections.

\begin{algorithm}[htb]
\label{alg:workflow}
\footnotesize
\caption{Dependency-Aware Multi-Robot Task Decomposition and Execution}
\begin{algorithmic}[1]
\REQUIRE Natural language instruction $\mathbb{I}$, environment information $\mathbb{E}$, robot set $\mathbb{R}$ with skill sets $\mathbb{S}$ and a few-shot example set $\mathbb{F}$
\STATE \textbf{Instruction Parsing}
\STATE Provide QA LLM module with $\mathbb{I}$ for parsing
\STATE Generate prompt $\mathbb{P} = (\mathbb{I}, \mathbb{E}, \mathbb{R}, \mathbb{S}, \mathbb{F})$ with explicit instruction that robot skills can combine through teamwork
\STATE Obtain parsed command structure in JSON format, including subtasks, robot assignments, dependencies, and object keywords
\STATE \textbf{Dependency-Aware Multi-Robot Task Decomposition}
\STATE Decompose $\mathbb{I}$ into subtasks $\mathbb{T} = \{T^{1}_{t_1}, T^{1}_{t_2}, \ldots, T^{K}_{t_j}\}$ with QA LLM, where each $T^{k}_{t_j}$ corresponds to an atomic skill in $\mathbb{S}$
\STATE Define dependency relationships $\mathbb{D} \subseteq \mathbb{T} \times \mathbb{T}$ during decomposition
\STATE Represent dependencies using Directed Acyclic Graph $G = (\mathbb{T}, \mathbb{D})$, where $\mathbb{D}$ enables task coordination and implicit team formation
\STATE \textbf{Dependency-Aware Task Parsing and Robot Assignment}
\STATE Initialize object map database with object locations and attributes using VLM-based object detector
\FOR{each subtask $T^{k}_{t_j}$ in topological order of DAG $G$}
    \IF{$T^{k}_{t_j}$ has no dependencies in $\mathbb{D}$}
        \STATE Execute $T^{k}_{t_j}$ in parallel
    \ELSE
        \STATE Wait for all dependencies of $T^{k}_{t_j}$ in $\mathbb{D}$ to complete
        \STATE Execute $T^{k}_{t_j}$
    \ENDIF
\ENDFOR
\FOR{each subtask $T^{k}_{t_j} \in \mathbb{T}$}
    \IF{QA LLM specified a particular robot for $T^{k}_{t_j}$}
        \STATE Assign $T^{k}_{t_j}$ to the specified robot from $\mathbb{R}$ with corresponding skill in $\mathbb{S}$
    \ELSE
        \STATE Use Breakdown function parser to find available robot of specified type from $\mathbb{R}$ with matching skill
    \ENDIF
\ENDFOR
\STATE \textbf{Task Execution using Actuation Module}
\FOR{each parsed subtask $T^{k}_{t_j} \in \mathbb{T}$}
    \STATE Execute assigned robot-specific atomic skill based on $\mathbb{S}$
\ENDFOR
\STATE \textbf{Update Object Map Database}
\STATE Continuously update object map based on real-time VLM object detector feedback
\end{algorithmic}
\end{algorithm}

\subsection{Instruction Parsing and Dependency-Aware Multi-Robot Task Decomposition Using the QA LLM Module}
\label{subsec: Instruction Parsing and Task Decomposition use the QA LLM Module}
This subsection details steps 1-8 of Algorithm 1, focusing on how high-level natural language instructions \( \mathbb{I} \) are processed by the QA LLM module and decomposed into executable subtasks with dependency relationships.

High-level natural language instructions \( \mathbb{I} \) serve as the input for DART-LLM. The input is first processed by the QA LLM for Instruction Parsing and Dependency-Aware Task Decomposition. The QA LLM parses the high-level instructions \( \mathbb{I} \) while considering the environment \( \mathbb{E} \) and the skills \( \mathbb{S} \) possessed by all robots \( \mathbb{R} \). 

In this module, we employ a DAG-based task decomposition system that decomposes a given instruction $\mathbb{I}$ into subtasks while also generating a dependency list for each subtask to ensure correct execution sequencing.  
As shown in Fig.~\ref{Fig:dag_example}, when instruction $\mathbb{I}$ is decomposed into a set of subtasks $T$ by QA LLM, each subtask \( T^{k}_{t_j} \) is assigned a dependency list $d_k$, which specifies all prerequisite subtasks that must be completed before executing \( T^{k}_{t_j} \). The system then determines the execution order $t_i$ automatically based on these dependencies and constructs a directed edge set $D_k$ for each $d_k$, thereby forming a subtask execution graph as illustrated. Finally, the system executes subtasks strictly in the specified order until the overall task objective is achieved.

During Task Decomposition, DART-LLM uses awareness of dependency relationships \( \mathbb{D} \) to decompose the task into multiple subtasks \( \mathbb{T} = \{T_{t_1}^1, T_{t_2}^1, \ldots, T_{t_j}^K \} \), where each subtask corresponds to an atomic skill in \( \mathbb{S} \). This facilitates parallel execution of decomposed subtasks without mutual dependencies and clearly expresses dependencies between subtasks.

Therefore, the prompt \( \mathbb{P} \) for the QA LLM is defined as 
\(
\mathbb{P} = (\mathbb{I}, \mathbb{E}, \mathbb{R}, \mathbb{S}, \mathbb{F})
\)
, where \( \mathbb{F} \) represents the few-shot example, with explicit instructions that robot skills can combine through teamwork.

The output of the QA LLM uses a structured standard format to handle the parsed commands and their dependencies. 
The standardized structure follows JSON syntax, as shown below:

\begin{lstlisting}[style=jsonstyle]
{
    "instruction_function": {
        "name": "<breakdown function 1>",
        "dependencies":["<dep 1>", "<dep 2>", …, "<dep n>"]
    },
    "object_keywords": ["<key 1>", "<key 2>", …, "<key n>"],

    "instruction_function": {
        "name": "<breakdown function 2>",
        "dependencies":["<dep 1>", "<dep 2>", …, "<dep n>"]
    },
    "object_keywords": ["<key 1>", "<key 2>", …, "<key n>"],

    ...
    
    "instruction_function": {
        "name": "<breakdown function m>",
        "dependencies":["<dep 1>", "<dep 2>", …, "<dep n>"]
    },
    "object_keywords": ["<key 1>", "<key 2>", …, "<key n>"]
}
\end{lstlisting}

The structured standard format message contains the names of the breakdown functions (i.e., atomic action skills) along with their parameters, object keywords, and dependencies.
These atomic action skills belong to \(\mathbb{S}\). For specific atomic action skills, please refer to subsection \ref{subsec:Grounding}. The parameters of the atomic action skills are searched within the \textit{object map} database. For details, please refer to subsection \ref{subsec:Object_Detector_Module}.

\subsection{Dependency-Aware Task Parsing and Robot Assignment Using the Breakdown Function Modules}
\label{subsec:Breakdown_Function_Modules}
This subsection elaborates on the implementation details of steps 9-25 of Algorithm 1, particularly how the system processes tasks according to their dependencies and assigns robots to subtasks.

The system first initializes the object map database with object locations and attributes using the VLM-based object detector. This provides the environmental context necessary for task execution and robot coordination.

Following the DAG structure established during task decomposition, the system processes each subtask $T^{k}_{t_j}$ in topological order of the graph $G$. This ordering ensures that subtasks are executed according to their dependencies:
- Subtasks without dependencies are executed in parallel, maximizing efficiency
- Subtasks with dependencies wait for their prerequisite tasks to complete before execution

The \textit{Breakdown Function Parser} module then processes each subtask $T^{k}_{t_j} \in \mathbb{T}$ for robot assignment. If the QA LLM has specified a particular robot for a subtask, that robot is directly assigned to the task. Otherwise, the \textit{Breakdown Function Parser} finds an available robot of the specified type from $\mathbb{R}$ with the matching skill required for the subtask.

This dependency-aware approach to task parsing and robot assignment ensures that complex tasks requiring collaboration between multiple robots are executed efficiently while respecting the necessary order of operations defined by the dependency relationships in $\mathbb{D}$.

\subsection{Task Execution and Grounding Using the Actuation Module}
\label{subsec:Grounding}
This subsection corresponds to steps 26-29 of Algorithm 1, following the dependency-aware execution and robot assignment described previously, detailing how the system grounds instructions in robot actions through the actuation module.

The \textit{Actuation} module grounds the instructions in embodiments by executing each parsed subtask $T^{k}_{t_j} \in \mathbb{T}$ using the corresponding atomic skill from \(\mathbb{S}\). All atomic action skills are asynchronously executed by the \textit{Actuation} module.

These atomic action skills are divided into two categories: navigation skills and robot-specific skills, both contained within \(\mathbb{S}\). The implementation of navigation skills is achieved through the ROS Navigation stack \cite{macenski2020marathon2}, while robot-specific skills are realized by creating specific action sets for each robot. 

\subsubsection{Navigation Skills}
The navigation skills are responsible for directing the movement and area access permissions of all robots or specific robots within a designated area. These skills include whether to avoid certain areas, return to the initial position, or move to a specified location. Table ~\ref{tab:description_navigation_functions} provides detailed descriptions of each navigation skill.

\begin{table*}[t]
\centering
\caption{Navigation Skill Descriptions}
\label{tab:description_navigation_functions}
\begin{tabular}{cll}
\hline
\textbf{Number} & \textbf{Skill Name} & \textbf{Description} \\
\hline
N1 & \texttt{avoid\_areas\_for\_all\_robots} & Sets the cost map to make all robots avoid specified areas. \\
   & \texttt{avoid\_areas\_for\_specific\_robots} & Sets the cost map for selected robots to avoid specified areas. \\
N2 & \texttt{target\_area\_for\_all\_robots} & Guides all robots to target points near the specified area. \\
   & \texttt{target\_area\_for\_specific\_robots} & Guides selected robots to target points near the specified area. \\
N3 & \texttt{allow\_areas\_for\_all\_robots} & Configures the cost map to allow all robots to access specified areas. \\
   & \texttt{allow\_areas\_for\_specific\_robots} & Configures the cost map for selected robots to access specified areas. \\
N4 & \texttt{return\_to\_start\_for\_all\_robots} & Instructs all robots to return to their initial starting position. \\
   & \texttt{return\_to\_start\_for\_specific\_robots} & Instructs selected robots to return to their initial starting position. \\
\hline
\end{tabular}
\end{table*}

\subsubsection{Robot Skills}
The robot-specific skills are tailored to the unique operational capabilities of different robot types, such as excavators and dump trucks. These skills handle tasks such as digging, unloading, and loading. Table ~\ref{tab:description_robot_functions} provides detailed descriptions of each robot skill.
It should be noted that different types of robots have different atomic action skill definitions.

\begin{table*}[t]
\centering
\caption{Robot Skill Descriptions}
\label{tab:description_robot_functions}
\begin{tabular}{cll}
\hline
\textbf{Number} & \textbf{Skill Name} & \textbf{Description} \\
\hline
FE1 & \texttt{excavator\_digging} & Instructs the excavator to dig at the specified target location. \\
FE2 & \texttt{excavator\_unloading} & Instructs the excavator to unload at the specified target location. \\
FD1 & \texttt{dump\_loading} & Instructs the dump truck to load materials at the specified target location. \\
FD2 & \texttt{dump\_unloading} & Instructs the dump truck to unload materials at the specified target location. \\
\hline
\end{tabular}
\end{table*}

\subsection{Update Object Map Database Using the VLM-based Object Detector Module}
\label{subsec:Object_Detector_Module}
This subsection explains steps 30-31 of Algorithm 1, describing how the system continuously updates its object map database through real-time sensing, which is crucial for maintaining an accurate representation of the environment throughout task execution.

The DART-LLM updates the \textit{object map} database in environment \( \mathbb{E} \) using the VLM-based object detector module. The object detector focuses on identifying the object keywords that were extracted during QA LLM's instruction parsing phase, ensuring that all entities relevant to the current task are properly detected and tracked. 

This module first converts all images captured by Unmanned Aerial Vehicles (UAVs) into bird's-eye views. These views are processed to detect and recognize objects in the environment \( \mathbb{E} \) . Selective search \cite{felzenszwalb2004efficient} is then used on the bird's-eye views to generate candidate bounding boxes. Following region proposals, a CLIP-based model \cite{radford2021learning} is employed for identification, updating the \textit{object map} database with the names, locations, and shapes of the objects. The updated information is then used by the Breakdown Function Handler to ensure accurate parameter resolution for the atomic skills during execution.

\section{Experiments}
\label{sec: Experiments}
\subsection{Experimental Setup}
To evaluate the performance of the DART-LLM system across different task sets and conduct a quantitative comparison with baseline methods, this study employs construction robots that are highly sensitive to subtask sequencing. These robots serve as test subjects to evaluate the system's effectiveness. Errors in subtask sequencing can lead to task failure. For instance, if a dump truck departs from the soil pile before the excavator has completed unloading, the loading and unloading operation will be unsuccessful.
We constructed a benchmark dataset designed specifically for evaluating natural language-driven task planning in multi-robot construction scenarios. The dataset includes three task levels: L1, L2, and L3. The task numbering reflects their respective levels.
L1 tasks are fundamental tasks involving a single robot equipped with all necessary skills, eliminating the need for multi-robot coordination.
L2 tasks require collaboration among multiple robots. These tasks consist of subtasks that must be executed in a specific sequence, though each subtask can still be completed independently by a single robot.
L3 tasks are complex, requiring multiple robots to work together. These tasks are decomposed into multiple interdependent subtasks, which must follow a strict execution order but may allow some parallel execution where feasible.
The dataset comprises 102 high-level instructions: 47 L1 tasks, 33 L2 tasks, and 22 L3 tasks. It is available on our project website: \href{https://wyd0817.github.io/project-dart-llm/}{project-dart-llm}.
Task execution can be visually validated in both simulated and physical robot environments. The simulation environment is built on the Unity platform and employs the PhysX physics engine. Both the simulation and physical environments feature two C30R tracked transport robots manufactured by Yanmar (Japan) and one ZX120 excavator manufactured by Hitachi Construction Machinery (Japan).
Table~\ref{tab:task_description} presents two specific example tasks for each task level, along with the required skills and applicable robot types.

\begin{table*}[!t]
\centering
\caption{Description of specific tasks in each task level}
\label{tab:task_description}
\resizebox{\textwidth}{!}{
\begin{tabular}{cclll}
\hline
\textbf{Task Level} & \textbf{Task Number} & \textbf{Task Description} & \textbf{Skill Number} & \textbf{Applicable Robots}\\
\hline
\multirow{2}{*}{Level 1} & L1-T1-001 & Inspect a puddle & N1, N2, N3, N4 & Excavator or Dump Truck\\
& L1-T2-001 & Clear an obstacle & N1, N2, N3, N4, FE1, FE2 & Excavator\\
\hline
\multirow{2}{*}{Level 2} & L2-T1-001 & Excavate soil & N1, N2, N3, N4, FE1, FD1, FE2, FD2 & Excavator, Dump Truck\\
& L2-T2-001 & Transport soil to the dump truck's initial position & N1, N2, N3, N4, FE1, FD1, FE2, FD2 & Excavator, Dump Truck\\
\hline
\multirow{2}{*}{Level 3} & L3-T1-001 & Clear the obstacle, then dig soil & N1, N2, N3, N4, FE1, FD1, FE2, FD2 & Excavator, Dump Truck\\
& L3-T2-001 & Clear the obstacle, then inspect the puddle & N1, N2, N3, N4, FE1, FE2, FE2, FD2 & Excavator, Dump Truck\\
\hline
\end{tabular}
}
\end{table*}

\subsection{Evaluation Metrics}

To evaluate the performance of the proposed approach on the dataset, this study employs five evaluation metrics:

\begin{itemize}
\item \textbf{SR (Success Rate)}: Whether the task is completely successful.

\item \textbf{IPA (Instruction Parsing Accuracy)}: The accuracy of instruction parsing, defined as the proportion of correctly parsed instructions mapped to the correct atomic action function names and parameters.

\item \textbf{DSR (Dependency Satisfaction Rate)}: The rate of dependency satisfaction, referring to the proportion of subtasks completed in the correct order of dependencies.

\item \textbf{SGSR (Semantic Grounding Success Rate)}: The rate of successful semantic grounding, defined as the proportion of code generated that can be successfully parsed by the breakdown function parse module. 

\item \textbf{RTR (Response Time Reliability)}: Assesses the model’s response time stability across multiple executions, using average response time and standard deviation to evaluate performance consistency under different conditions.
\end{itemize}
The evaluation relies on dataset ground truth to measure each metric.

\subsection{Experimental Results}
\subsubsection{Evaluation Against Baselines Using Different LLMs}
This study evaluated DART-LLM using multiple foundation models: Llama-3.1-8B \cite{dubey2024llama}, GPT-4o \cite{hurst2024gpt}, GPT-3.5-turbo \cite{openai_gpt3.5turbo}, Claude-3.5-Haiku \cite{anthropic_haiku}, and DeepSeek-r1-671B \cite{guo2025deepseek}. We also compared our approach with the SMART-LLM baseline implemented with Llama-3.1-8B and DeepSeek-r1-671B. Table~\ref{tab:results} shows the evaluation results across tasks of L1, L2, and L3 complexity levels. 
Notably, all 102 test cases were unseen data, using few-shot examples solely for prompting and excluding them from the test set.
All evaluation metrics range from 0 to 1, with higher values indicating better performance.

The results indicate that for task level L1, all implementations achieved perfect scores across SR, IPA, DSR, and SGSR metrics (all 1.00). However, notable differences appear in the RTR metric, where DART-LLM with Llama3.1 achieved the highest score of 0.96, significantly outperforming other models including SMART-LLM implementations (0.24 and 0.08).

As task complexity increases to L2, performance differences become more pronounced. DART-LLM with DeepSeek-r1 achieved the highest SR of 0.97. GPT-4o followed with 0.96, Claude3.5 with 0.90, GPT-3.5-turbo with 0.87, and Llama3.1 with 0.85. All DART-LLM implementations maintained perfect IPA scores, demonstrating the effectiveness of proposed structured JSON format and dependency-aware task decomposition method. In contrast, the SMART-LLM baseline showed substantial performance degradation, with SR scores of 0.36 and 0.78 for Llama3.1 and DeepSeek-r1 respectively, and reduced IPA scores (0.57 and 0.78).
The explicit specification of dependencies in proposed approach significantly improved model logical reasoning capabilities, as evidenced by the high DSR scores across all DART-LLM implementations compared to the SMART-LLM baseline. For RTR, DART-LLM with Llama3.1 again demonstrated superior performance (0.90) compared to other models, while SMART-LLM implementations struggled significantly (0.19 and 0.04).

At task level L3, DART-LLM with DeepSeek-r1 demonstrated the best performance with an SR of 0.94. GPT-4o followed with 0.93, while Claude3.5, Llama3.1, and GPT-3.5-turbo achieved 0.86, 0.84, and 0.80 respectively. The SMART-LLM baseline performance degraded substantially, with SR scores dropping to 0.24 and 0.65 for Llama3.1 and DeepSeek-r1 respectively.

The structured JSON format consistently contributed to high RTR scores for DART-LLM (Llama3.1), maintaining 0.96, 0.90, and 0.86 across L1, L2, and L3 respectively. In contrast, the SMART-LLM baseline implementations showed markedly lower RTR scores (ranging from 0.02 to 0.24), primarily due to its direct generation of more complex Python code, which led to increased planning time for code generation. While the SMART-LLM baseline achieved reasonable SR scores with DeepSeek-r1, its low RTR indicates significant practical limitations in real-world applications where code execution is essential.

Overall, the DART-LLM system showed superior performance when using DeepSeek-r1 for SR metrics and Llama3.1 for RTR metrics. Notably, the Llama3.1 model, with only 8B parameters, performed competitively with much larger models and achieved significantly higher RTR scores than all other approaches. This suggests that using smaller models with proposed JSON-based command generation approach provides an excellent balance between performance and deployability, offering better real-time capabilities while maintaining high success rates.

\begin{table*}[tb]
\centering
\small
\caption{Evaluation of DART-LLM and SMART-LLM baseline across L1, L2, and L3 tasks.}
\label{tab:results}
\resizebox{1.0\textwidth}{!}{
\begin{tabular}{lcccccccccccccccccccccccc}
\hline
\multirow{4}{*}{\textbf{Models}} &  & \multicolumn{5}{c}{\multirow{2}{*}{\textbf{L1}}} &  & \multicolumn{5}{c}{\multirow{2}{*}{\textbf{L2}}} &  & \multicolumn{5}{c}{\multirow{2}{*}{\textbf{L3}}} \\ &  & \multicolumn{5}{c}{} &  & \multicolumn{5}{c}{} &  & \multicolumn{5}{c}{} \\ 
\cline{3-7}\cline{9-13}\cline{15-19}
&  & \multirow{2}{*}{\textbf{SR}} & \multirow{2}{*}{\textbf{IPA}} & \multirow{2}{*}{\textbf{DSR}} & \multirow{2}{*}{\textbf{SGSR}} & \multirow{2}{*}{\textbf{RTR}} &  & \multirow{2}{*}{\textbf{SR}} & \multirow{2}{*}{\textbf{IPA}} & \multirow{2}{*}{\textbf{DSR}} & \multirow{2}{*}{\textbf{SGSR}} & \multirow{2}{*}{\textbf{RTR}} &  & \multirow{2}{*}{\textbf{SR}} & \multirow{2}{*}{\textbf{IPA}} & \multirow{2}{*}{\textbf{DSR}} & \multirow{2}{*}{\textbf{SGSR}} & \multirow{2}{*}{\textbf{RTR}} \\ &  &  &  &  &  &  &  &  &  &  &  &  &  &  &  &  &  &  \\ 
\hline
&  &  &  &  &  &  &  &  &  &  &  &  &  &  &  &  &  &  \\

\begin{tabular}[l]{@{}c@{}}{DART-LLM~}{(Llama3.1)}\end{tabular} &  & \textbf{1.00} & \textbf{1.00} & \textbf{1.00} & \textbf{1.00} & \textbf{0.96} &  & 0.85  & \textbf{1.00} & 0.85 & 0.90 & \textbf{0.90} &  & 0.84 & 0.93 & 0.84 & 0.88 & \textbf{0.86}  \\ &  &  &  &  &  &  &  &  &  &  &  &  &  &  &  &  &  &  \\
\begin{tabular}[l]{@{}c@{}}{DART-LLM~}{(GPT-3.5-turbo)}\end{tabular} &  & \textbf{1.00} & \textbf{1.00} & \textbf{1.00} & \textbf{1.00} & 0.75 &  & 0.87  & \textbf{1.00} & 0.87 & 0.87 & 0.68 &  & 0.80 & 0.97 & 0.85 & 0.80 & 0.66 \\ &  &  &  &  &  &  &  &  &  &  &  &  &  &  &  &  &  &  \\
\begin{tabular}[l]{@{}c@{}}{DART-LLM~}{(GPT-4o)}\end{tabular} &  & \textbf{1.00} & \textbf{1.00} & \textbf{1.00} & \textbf{1.00} & 0.55 &  & 0.96  & \textbf{1.00} & 0.96 & 0.96 & 0.50 &  & 0.93  & \textbf{1.00} & 0.93 & 0.93 & 0.45 \\ &  &  &  &  &  &  &  &  &  &  &  &  &  &  &  &  &  &  \\
\begin{tabular}[l]{@{}c@{}}{DART-LLM~}{(Claude3.5)}\end{tabular} &  & \textbf{1.00} & \textbf{1.00} & \textbf{1.00} & \textbf{1.00} & 0.70 &  & 0.90 & \textbf{1.00} & 0.90 & 0.95 & 0.63 &  & 0.86 & \textbf{1.00}  & 0.86  & 0.90 & 0.60 \\ &  &  &  &  &  &  &  &  &  &  &  &  &  &  &  &  &  &  \\
\begin{tabular}[l]{@{}c@{}}{DART-LLM~}{(DeepSeek-r1)}\end{tabular} &  & \textbf{1.00} & \textbf{1.00} & \textbf{1.00} & \textbf{1.00} & 0.30 &  & \textbf{0.97} & \textbf{1.00} & \textbf{0.97} & \textbf{0.97} & 0.25 &  & \textbf{0.94} & \textbf{1.00}  & \textbf{0.94}  & \textbf{0.94} & 0.20 \\ &  &  &  &  &  &  &  &  &  &  &  &  &  &  &  &  &  &  \\
\begin{tabular}[l]{@{}c@{}}{SMART-LLM}{(Llama3.1)~\cite{kannan2023smart}}\end{tabular} &  & \textbf{1.00} & \textbf{1.00} & \textbf{1.00} & \textbf{1.00} & 0.24 &  & 0.36 & 0.57 & 0.46 & 0.36 & 0.19 &  & 0.24& 0.37  & 0.43  & 0.24 & 0.16 \\ &  &  &  &  &  &  &  &  &  &  &  &  &  &  &  &  &  &  \\
\begin{tabular}[l]{@{}c@{}}{SMART-LLM}{(DeepSeek-r1)~\cite{kannan2023smart}}\end{tabular} &  & \textbf{1.00} & \textbf{1.00} & \textbf{1.00} & \textbf{1.00} & 0.08 &  & 0.78 & 0.78 & 0.91 & 0.87 & 0.04 &  & 0.65 & 0.65  & 0.86  & 0.78 & 0.02 \\ &  &  &  &  &  &  &  &  &  &  &  &  &  &  &  &  &  &  \\
\hline
\end{tabular}}
\vspace{-1mm}
\end{table*}

\subsubsection{Ablation Study}
To evaluate the effectiveness of proposed dependency-aware approach, we conducted an ablation study on L3 complexity tasks using five different LLMs. We compared two conditions: "With Dependency" (i.e., using a DAG to represent task dependencies) and "Without Dependency" (i.e., not using a DAG to specify task relationships).

\begin{figure}[t]
\centering
\includegraphics[width=\linewidth]{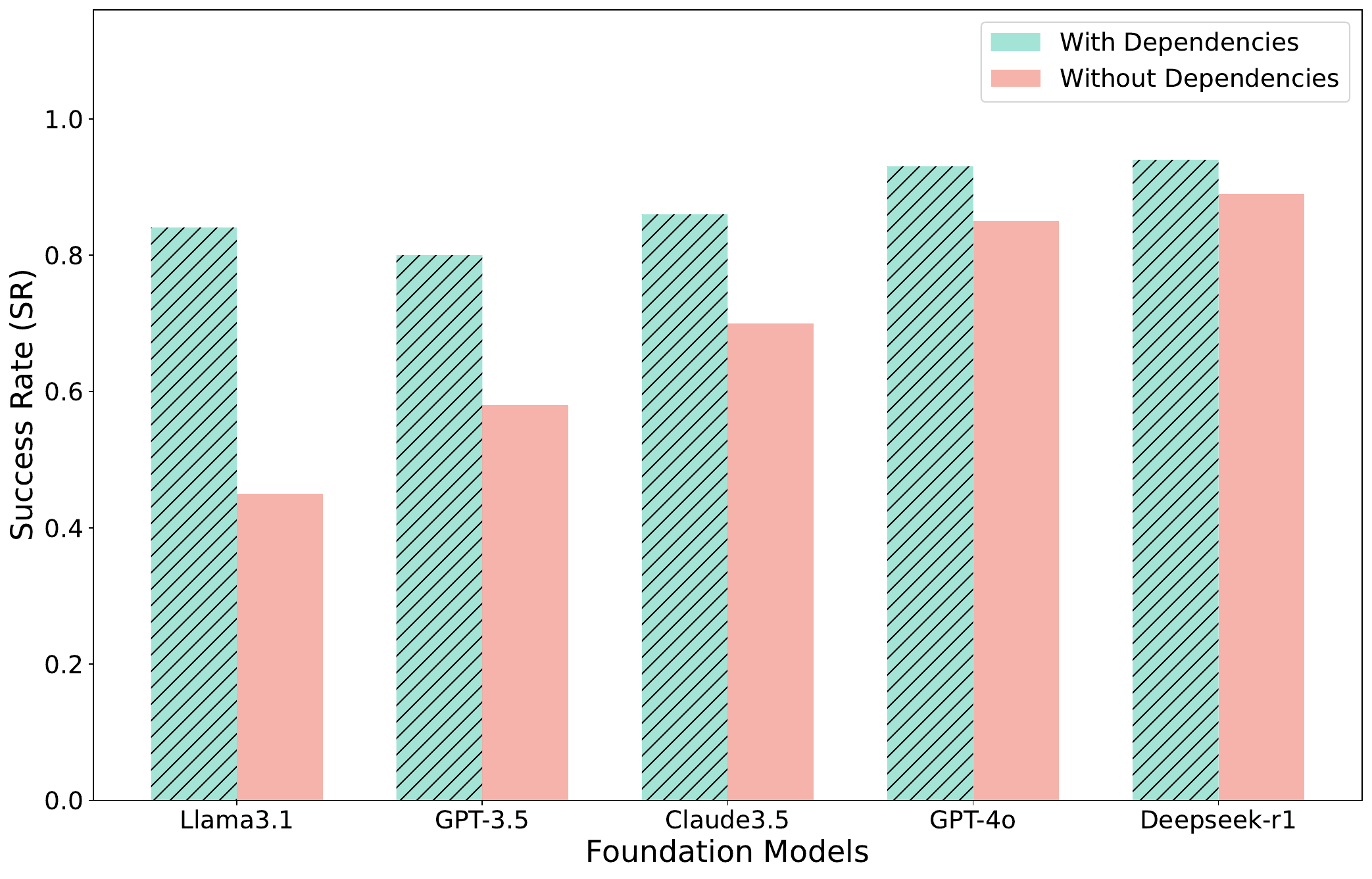}
\caption{Success Rate (SR) comparison between "With Dependencies" (using DAG) and "Without Dependencies" (not using DAG) for L3 complexity tasks across different models.}
\label{fig:Ablation_Comparison}
\end{figure}

Fig.~\ref{fig:Ablation_Comparison} presents the SR under both conditions. The experimental results indicate that the "Without Dependency" condition (i.e., without using a DAG) significantly reduces the performance of all models. Among them, Llama3.1 exhibited the largest performance drop (from 0.84 to 0.45). Even state-of-the-art models such as GPT-4o and DeepSeek-r1 experienced notable declines (from 0.93 to 0.85 and from 0.94 to 0.89, respectively). The pronounced performance gap in smaller models suggests that explicit dependency modeling via DAG can effectively compensate for their limited reasoning capabilities, demonstrating that proposed approach enables smaller, more deployable models to achieve competitive performance on complex tasks.

\subsubsection{Deployment in Real World}
\begin{figure}[thpb]
\centering
\includegraphics[width=\linewidth]{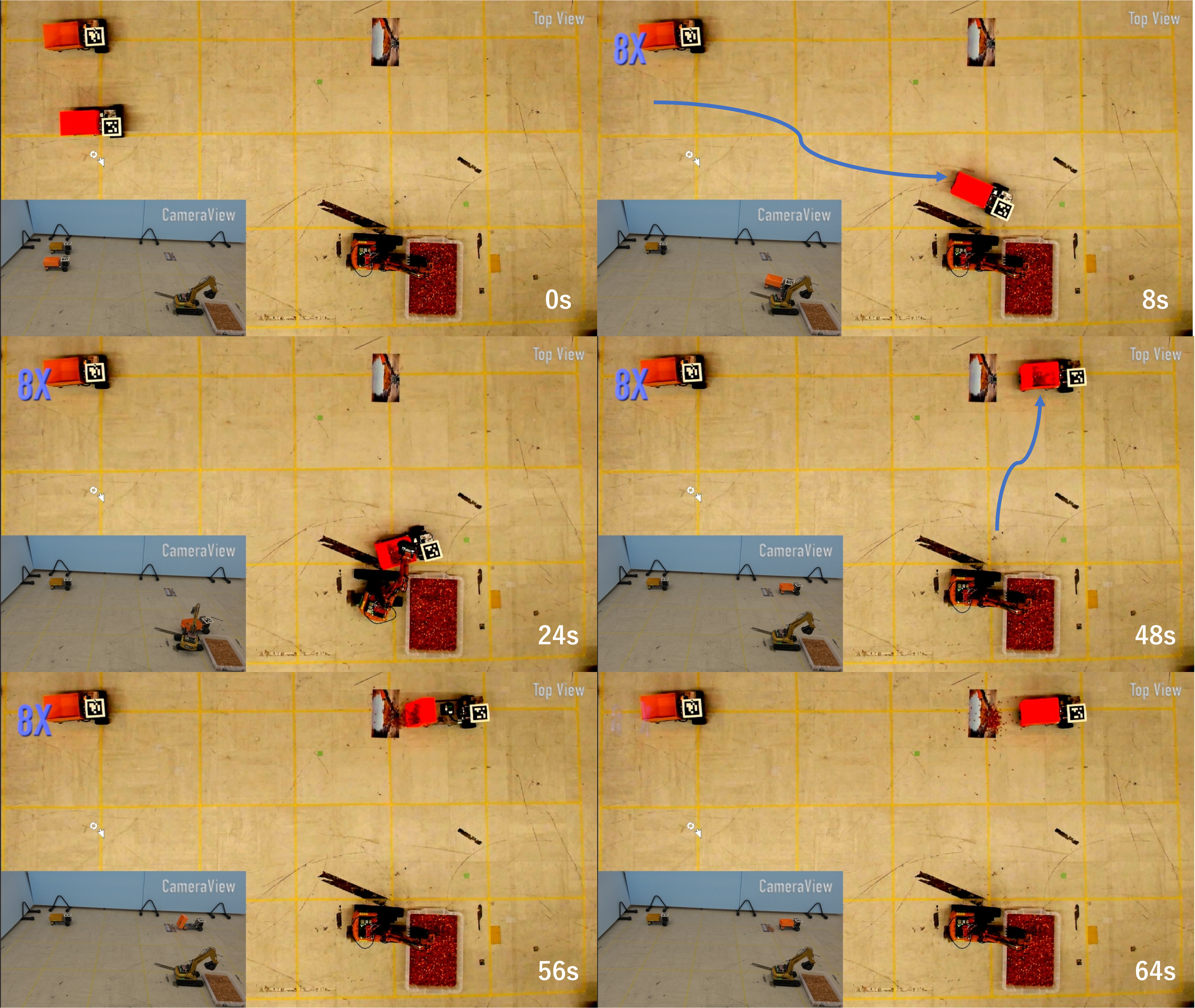}
\caption{DART-LLM (Llama3.1) in the L2-T1-001 Task Using Real Robots: The sequence begins at 0 [s] with the initial positioning of the Dump Truck at the starting location. At 8 [s], the Excavator prepares to dig soil. By 24 [s], the Excavator transfers the excavated soil into the Dump Truck, completing the loading operation. At 48 [s], the loaded Dump Truck navigates towards the designated dumping area. At 56 [s], the Dump Truck unloads the soil at the puddle location. Finally, at 64 [s], the Dump Truck completes the task by retracting its vessel.}
\label{fig:L2T1_REAL}
\end{figure}

To evaluate the deployment capability of DART-LLM, we conducted tests in a real-world environment. Fig.~\ref{fig:L2T1_REAL} presents the results of the L2-T1-001 task using the Llama3.1 model. The results indicate that DART-LLM demonstrates good applicability on real robots.

For more experiments, please visit the project website \href{https://wyd0817.github.io/project-dart-llm/}{project-dart-llm}.

\section{Conclusion and Future Work}
\label{sec: Conclusion_and_Future_Work}
This paper presented DART-LLM, a novel framework that leverages LLMs and DAG-based dependency modeling for coordinated multi-robot task execution. The proposed system integrates four key modules: a QA LLM module for instruction parsing and dependency-aware task decomposition, a Breakdown Function module for task parsing and robot assignment, an Actuation module for executing robot-specific skills, and a VLM-based object detector module for environmental perception. 
This architecture demonstrated superior performance compared to existing methods across all evaluation metrics, with DART-LLM (DeepSeek-r1) achieving a 94\% success rate on complex tasks while maintaining perfect instruction parsing accuracy. The smaller Llama3.1 model exhibited exceptional response time reliability, making it suitable for resource-constrained deployments. Furthermore, ablation experiments reveal that DAG-based dependency modeling significantly enhances model performance. Notably, it compensates for the limited reasoning capabilities of smaller models, enabling them to execute complex tasks more effectively. 
Real-world testing validated the system's practical applicability. 
Future work will focus on scaling DART-LLM for larger multi-robot teams to enhance system applicability. Additionally, optimizing the trade-off between model size and performance will be investigated to accommodate diverse deployment scenarios.

\section*{ACKNOWLEDGMENT}
This work is supported by JST [Moonshot Research and Development], Grant Number [JPMJMS2032]. The work was first submitted to an IEEE conference on September 15, 2024.


\end{document}